\def\year{2019}\relax
\documentclass{article} 
\usepackage{aaai18}  
\usepackage{times}  
\usepackage{helvet}  
\usepackage{courier}  
\usepackage{amssymb,amsmath,amsthm}
\usepackage{url}  
\usepackage{graphicx}  
\graphicspath{ {./images/} }
\usepackage[utf8]{inputenc}  
\usepackage[english]{babel}  
\usepackage[linesnumbered,ruled,vlined]{algorithm2e}
\usepackage{biblatex}
\newtheorem{definition}{Definition}
\newtheorem{proposition}{Proposition}
\newtheorem{lemma}{Lemma}

\newtheorem{assumption}{Assumption}
\newtheorem{theorem}{Theorem}
\newtheorem{corollary}{Corollary}
\newtheorem{observation}{Observation}

\newcommand{\nonlin}{\mathcal{P}_{\!\!r}}
\newcommand{\bottom}{\perp}
\newcommand{\goods}{\mathcal{G}}
\newcommand{\policies}{\Pi}
\newcommand{\mdps}{\mathcal{M}}
\newcommand{\prob}{\mathbb{P}}
\newcommand{\optimals}{\textnormal{Opt}}

\DeclareMathOperator*{\argmax}{arg\,max}

\begin{document}
%
\title{Introspection Learning}
\author{C. R. Serrano and Michael A. Warren\\
  HRL Laboratories, LLC \\
  Malibu\\
}
\maketitle
\begin{abstract}
  Traditional reinforcement learning agents learn from experience, past or 
  present, gained through interaction with their environment.  Our approach 
  synthesizes experience, without requiring an agent to interact with their 
  environment, by asking the policy directly ``Are there situations $X$, $Y$, 
  and $Z$, such that in these situations you would select actions $A$, $B$, 
  and $C$?'' In this paper we present Introspection Learning, an algorithm 
  that allows for the asking of these types of questions of neural network 
  policies. Introspection Learning is reinforcement learning  algorithm agnostic and the states 
  returned may be used as an indicator of the health of the policy or to 
  shape the policy in a myriad of ways. We demonstrate the usefulness
  of this algorithm both in the context of speeding up training and
  improving robustness with respect to safety constraints.
\end{abstract}

\noindent

\section{Introduction}

One notable feature of human learners is that we are able to
carry out counter-factual reasoning over unrealized events.  That is,
we contemplate potential answers to questions of the form, ``What
would I do in situations $X$, $Y$, and $Z$?''  A related, and perhaps
more pertinent, form of question is, ``Are there situations $X$, $Y$, and $Z$,
such that in these situations I would select actions $A$, $B$, and
$C$?''  In this case, the actions $A$, $B$, etc., might be
actions that are likely to result in particularly good or bad
outcomes, and answers $X$, $Y$, etc., can be useful, especially when
they are of an unexpected nature, since they reveal potential failures
of robustness (in the case of bad examples) or potential strengths (in
the case of good examples).  In this paper, we describe a novel
approach to answering and utilizing the answers to questions of this
form when asked not of a human agent, but of a reinforcement learning
agent.  Our approach is not based solely on the deployment of
techniques from the typical machine learning toolbox, as we make
crucial use of SMT-solving, which is more familiar to researchers in
the field of formal methods.  In the theoretical development, we
capture our use of SMT-solving technology via the abstraction of what
we are calling \emph{introspection oracles}: oracles that may give us
direct access to sets of (state, action) pairs satisfying fixed
constraints with respect to the policy network.

By querying the oracle during training it is possible to generate
(state, action)-pairs capturing failures/strengths of the agent with
respect to properties of interest.  For instance, if there are certain
``obviously wrong'' actions that the agent should never take (e.g.,
selecting a steering angle that would cause the automobile controlled
by the policy network to drive off of the road when there are no
obstacles or other dangers present), we query the oracle as to whether
there exists states in which the agent would select such actions.  Our
algorithm then uses this data to train so as to improve the safety of
the agent and without requiring that such potentially dangerous or
costly situations be encountered in real life.  It is true that such
(state, action) pairs are potentially discoverable in
simulation/testing, but when the set of such pairs is known beforehand
we save time and improve policy robustness by generating them analytically.

In this paper, we introduce a new algorithm for reinforcement learning, which we
call the \emph{Introspection Learning Algorithm}, that exploits
introspection oracles to improve the training and robustness of reinforcement learning (RL) agents versus baseline training algorithms.  This algorithm
involves modifying the underlying MDP structure and we derive
theoretical results that justify these modifications.  Finally, we
discuss several experimental results that clearly showcase the
benefits to both performance and robustness of this approach.  In
particular, in the case of robustness, we evaluated our results by
querying the weights after training to determine numbers of Sat
(examples found), Unsat (examples mathematically impossible) and
Timeout (ran out of time to find or refute existence of examples)
results.

The paper is organized as follows.  In Section~\ref{oracles} we
introduce the mathematical abstraction of introspection oracles and
discuss briefly their embodiment as SMT-solvers.  Section
\ref{introspection} details our Introspection Learning
Algorithm.  Finally, Section \ref{methods} captures our
empirical results.  The Appendix (Section \ref{appendix}) includes the proof of
a basic result that justifies the modification of MDPs made in our algorithms.

\subsection{Related work}

Previously, Linear Programming, which is itself is a constraint
solving technique, has been employed in reinforcement
learning to constrain the exploration space for the agent’s policy to
improve both the speed of convergence and the quality of the policy
converged to \cite{Amos2017} or as a replacement for
more traditional Dynamic Programming methods in Q-Learning to solve
for equilibria policies in zero-sum multi-agent Markov game MDPs
\cite{Littman,Greenwald}.  Previous work has also been done on 
incorporating Quadratic Program solvers to restrict agent exploration to ``safe'' 
trajectories by constraining the output of a neural network policy 
\cite{Pham2017,Amos2017}.  Introspection Learning is fundamentally 
different from these approaches as rather than restricting the action space, or 
replacing our Q function, we are instead shaping our agents in policy space by 
asking our policy for state batches where it would satisfy stated 
constraints, without needing the agent to actually experience these
states.

Exciting recent work on verification of neural networks
(e.g., \cite{Reluplex,Lomuscio}) is closely related the work described
here.  In addition to the similarity of the techniques, we are indeed
capturing verification results as a robustness measure (see
below).  One practical distinction is that we are using the dReal
solver \cite{dReal}, which is able to handle
networks with general non-linear activations, but as a trade-off (not
made in other SMT-solvers) admits the possibility of
``false-positive'' $\delta$-satisfiable instances.  In principle, our
algorithm can be used with any compatible combination of SMT-solvers and neural
network architectures.  

\section{Introspection Oracles}\label{oracles}

In order to set the appropriate theoretical stage, we will first
introduce some notation and terminology.
\begin{definition}\label{definition:pre_mdp}
  A \emph{pre-Markov decision process} (pre-MDP) consists of a set $S$ of
  \emph{states}, a set $A$ of \emph{actions}, and \emph{transition
    probabilities} $p(s,a,s')$ in $[0,1]$ for $s,s'\in S$ and $a\in A$
  such that $\sum_{s'}p(s,a,s')=1$.
\end{definition}
Intuitively, the value $p(s,a,s')$ is the
probability $\prob(s'|s,a)$  transitioning from state $s$ to state
$s'$ on taking action $a$.
\begin{definition}\label{definition:policy}
  Given a \emph{pre-Markov Decision Process} (pre-MDP) $D=(S,A,p)$, a \emph{policy for $D$} assigns to each state $s$
  a probability distribution $\pi(s)$ over the set $A$.
\end{definition}
A pre-MDP is called a MDP$\backslash$R in, e.g., \cite{Abbeel}.

Often we are concerned with cases where $A$
is finite and the policies $\pi$ under consideration are \emph{deterministic} in the sense that,
for each state $s$, $\pi(s)(a)=0$ for all but a single element $a$ of
$A$.  When $p(s,a,s')=1$ we write ${a\colon s\to s'}$.  Given a pre-MDP
$D$, we denote by $\policies(D)$ the set of all
policies for $D$.
\begin{definition}\label{def:mdp}
  A \emph{Markov decision process} (MDP) consists of a pre-MDP
  $(S,A,p)$ together with a \emph{reward} function ${r\colon
  S\times A\to \mathbb{R}}$ which is bounded, a subset $T\subseteq
S\times A$
  of \emph{terminal (state, action)-pairs}, and a new state $s_{t}$
  not in $S$ such that:
  \begin{itemize}
  \item For any $(s,a)\in T$, $a\colon s\to s_{t}$;
  \item For any $a\in A$, $a\colon s_{t}\to s_{t}$; and
  \item For any $a\in A$, $r(s_{t},a)=0$.
  \end{itemize}
\end{definition}
One non-standard feature of Definition \ref{def:mdp} is that we consider terminal
pairs $(s,a)\in S\times A$ rather than terminal states.  This will be
technically useful below.  We also follow \cite{SuttonBarto} in that
the provision of terminal pairs modifies the pre-MDP
structure in adding a dummy stable state $s_{t}$ to which all terminal
states canonically transition such that subsequent transitions from
$s_{t}$ have no reward.  This is a
technical convenience which streamlines some of the theory.

We denote by $\mdps(D)$ the set of all Markov decision processes over
the pre-MDP $D$ and by $\policies(D)$ the set of all policies over
$D$.  Given an MDP $M$ in $\mdps(D)$, we denote by $\optimals(M)$ the
subset of $\policies(D)$ consisting of those policies that are optimal
for $M$. In broad strokes, inverse reinforcement learning \cite{Ng} is
concerned with, given a policy $\pi$ in $\policies(D)$ (or, more
often, a set of its
trajectories), determining an element $M$ of $\mdps(D)$ such that
$\pi$ is in $\optimals(M)$.  We are concerned with a closely related
problem.

One difference between our approach and that of inverse reinforcement
learning is that instead of assuming access to a target policy
$\pi$ or its trajectories, we assume that we have access to certain
\emph{properties} that target policies \emph{ought} to have.  In the
simplest case, such a property is given by a subset of the set
$S\times A$ of (state, action) pairs.\footnote{In the more general
  case, the relevant properties should be (non-empty) subsets of space
  $(S\times A)^{*}$ of finite sequences of (state, action) pairs that
  are compatible with the underlying transition probabilities of $D$.
In this paper, we restrict attention to the more elementary notion.}
We refer to policies with the required properties as \emph{good}
policies.  There is considerable flexibility in the notion of goodness
here, but in many cases it will be associated with safety and
robustness.  E.g., a good policy for driving a car would not make
unexpected sharp turns when the road ahead is straight and clear of
obstacles.  Much of our focus is on these kinds of examples, but it is
worth emphasizing that goodness could instead be associated with
performance  rather than safety.

In order to make the problem tractable, it is necessary to restrict to
sufficiently well-behaved subsets of $S\times A$.  For us, the
well-behaved subsets are those definable in the first-order theory of
real arithmetic with common non-linear function symbols (e.g., $\sin$,
$\log$, $\max$, $\tanh$, etc.).\footnote{In the experimental
  results captured in this paper, we restricted further to
  semialgebraic subsets.  I.e., those describable as finite unions of
  sets defined by finitely many polynomial equations and inequations.}
Denote by $\nonlin(X)$ the set of all such
subsets of $X\subset\mathbb{R}^{n}$.  With this notation in place, we
arrive the definition of introspection oracle.
\begin{definition}\label{def:introspection_oracle}
  Given policy $\pi$ in $\policies(D)$, an \emph{introspection oracle
    for $\pi$} is a map $\omega_{\pi}\colon\nonlin(S\times
  A)\to\{\bottom\}+S$ such that if $\omega_{\pi}(U)\neq\bottom$, then
  $(\omega_{\pi}(U),\pi(\omega_{\pi}(U)))$ is in $U$.  An
  introspection oracle is \emph{non-trivial} when there exists $U$ in
  $\nonlin(S\times A)$ such that $\omega_{\pi}(U)\neq\bottom$.
\end{definition}
Intuitively, an introspection oracle $\omega_{\pi}$ for
$\pi$ attempts to answer questions of the form: ``Are there inputs
that give rise via $\pi$ to a (state, action) pair with property $U$?''
Here $\perp$ is an error signal which can be provided with several
possible semantics.  Here it is best understood as indicating
that the oracle was unable to find an element of $U$  in a reasonable amount
of time.

Before turning to describe our use of introspection oracles in
reinforcement learning, we observe that non-trivial introspection
oracles do indeed exist: 
\begin{observation}
  For policy functions $\pi$ definable in the language of first-order
  real arithmetic with non-linear function symbols ($\sin$, $\cos$,
  $\log$, $\tanh$, etc.) there exist non-trivial introspection oracles.
\end{observation}
The existence of such introspection oracles which are moreover
practically useful in the sense of returning outputs $\neq\perp$ in
a wide range of feasible cases is guaranteed by the $\delta$-decision
procedure of Gao, Avigad and Clarke \cite{Gao}, which is implemented
in the dReal non-linear SMT-solver.  The novelty of dReal is that it
overcomes the undecidability of real arithmetic with non-linear
function symbols by accepting a compromise: whereas unsatisfiable
(Unsat) results are genuine, satisfiable (Sat) results may be
false-positives.  Note that, unlike in many of the other applications
of SMT-solving to verification of neural networks such as
\cite{Reluplex,Lomuscio}, dReal is able to handle all common non-linear
activations.  In terms of our abstraction, spurious Sat results, which
are easily detected by a forward pass of the network, can be regarded
as instances where $\omega_{\pi}(U)=\bottom$. 

\section{The Introspection Learning Algorithm}\label{introspection}

We now describe the Introspection Learning Algorithm in detail,
starting with its inputs.  First, this algorithm assumes given an
off-policy reinforcement learning algorithm (OPRL) and corresponding policy
function $\pi$.  It is furthermore assumed that $\pi$ is describable
in the language of real arithmetic with non-linear function symbols.

Additionally assume given a family $(U_{i})_{i}$ of
subsets $U_{i}\in\nonlin(S\times A)$, which will be used when we query
the oracle $\omega$.  Having a sufficiently rich family
$(U_{i})_{i}$ will provide a mechanism
for generating more useful examples and the design
of these properties is one of the main engineering challenges
involved in utilizing the algorithm effectively. Pairs $(s,\pi(s))$
obtained from the oracle as $s=\omega(U_{i})$ are added to the OPRL agent's 
replay buffer.
\begin{algorithm}\label{algorithm}
	\DontPrintSemicolon 
        \KwData{Off-policy RL algorithm OPRL, policy function
          $\pi$, family of queries $(U_{i})_{i}$, a schedule
          $\sigma$, a reward cutoff $R$}
	Initialize OPRL policy $\pi$ with random weights $\vartheta$
        and replay buffer $D$\;
	\For{episode $e\in \{1, \ldots, M\}$} {
		Train OPRL as specified\;
		\If{moving average reward $< R$ and $e\in\sigma$} {
                  For each $i$, query
                  $\omega_{\pi}(U_{i})$
                  and add examples
                  $\omega_{\pi}(U_{i})\in S$ to $D$ as terminal
		}
	}
	\caption{{\sc Introspection Learning}}
	\label{algo:introspection}
\end{algorithm}

Finally, we assume given a schedule determining when during training to perform
queries and updates.  For simplicity in describing the algorithm we
assume that the schedule is controlled by two factors.  First, a
simple set $\sigma$ of training indices.  Second, a bound $R$ on
moving average reward such that once moving average reward is greater
than or equal to $R$ we no longer perform queries or updates on gathered
examples.

In summary, given the aforementioned inputs, the Introspection
Learning Algorithm \ref{algorithm} proceeds by training $\pi$ as usual
according to the OPRL except that, when episode indices $e$ in $\sigma$
are arrived at and the moving average reward remains below $R$, the
oracle is queried with the specified family of pairs, examples are
gathered (when possible) and inserted into the replay buffer as
terminal.

Mathematically, this algorithm effectively produces a modified MDP structure
$M^{\dagger}$ by altering the terminal pairs and the reward structure.  In the
Appendix (Section \ref{appendix}), we show (Theorem \ref{theorem:equiv}) that,
under reasonable hypotheses, the sets of optimal policies for the
original MDP $M$ and the modified MDP $M^{\dagger}$ coincide.

There are several parameters and variations of this algorithm
possible, of which we now mention several.  First, in some cases it
may be necessary or useful to post-process the gathered
state batches (e.g., to ensure sufficient balance/symmetry
properties).  Here consideration should be paid to the bias introduced by state 
batches which are in one sense ``on policy'' (if the agent were in a state 
returned by the SMT-solver it would have taken the specified action with high 
probability), but are not guaranteed to be ``on trajectory'' as we have no 
guarantee the state would be reachable by policy $\pi$.  In practice,
we have found such processing to be unnecessary provided that
suitable $(U_{i})_{i}$ are selected and a reasonable schedule is followed.

In addition to varying the schedule, it is also possible to consider
a range of options for the behavior of the replay buffer and
how to train on the examples contained therein.  We have found it to
usually be sufficient to train on these as terminal
states with high-negative or high-positive reward, however 
other approaches can also be considered. It should be noted that treating these 
states as terminal will alter the optimal policy, which may or may not be 
desired, and alternatively one could query the training environment with the 
state batches and specified actions to recover the reward signal and next state 
from the environment in order to reduce the change in the optimal policy.  Our 
intention was to take a na\"ive approach as we are interested in applications 
where acquiring experience is potentially risky or expensive.

\section{Experimental Environments and Results}\label{methods}

Our experiments were conducted with the
Double Deep Q Network algorithm DDQN \cite{VanHasselt2015} with Prioritized 
Experience Replay\cite{Schaul2015} and the OpenAI Gym 
``Lunar Lander'' environment \cite{OpenAI}, OpenAI Gym ``Cliff Walk'' environment 
\cite{OpenAI} and the DeepMind AI Safety Gridworld ``Absent Supervisor'' 
environment \cite{SafetyGridworlds}.  Prioritized Experience Replay augments the 
selection of experience tuples from the DDQN replay buffer by preferentially 
selecting experience with high TD error and simultaneously correcting for the 
bias this introduces by scaling the loss in the neural network update 
proportionally to the size of the TD error.

In the ``Lunar Lander'' environment the objective is to safely land a spacecraft 
on the surface of the moon by controlling four discrete actions for each of its 
three engines.
The state space is eight dimensional with six continuous variables representing 
location in two-dimensional cartesian-coordinates, linear velocity,
angle and angular velocity, and two boolean  variables indicating
whether or not contact is being made with the ground by each of the
lander's two legs. The reward signal positively reinforces movement
toward the landing pad, as well as bonus for making leg contact with the ground.
Negative reward is given for moving away from the
landing pad or losing contact with the ground.  The environment is
considered solved when the agent achieves a 100 episode moving
average reward of at least 200.

\begin{figure}[h]
	\centering
	\includegraphics[width=85mm]{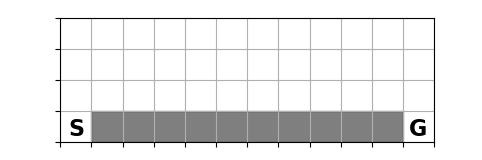}
	\caption{Cliff Walk Gridworld\label{fig:cliffwalk}}
\end{figure}

In the ``Cliff Walk'' gridworld environment (Figure \ref{fig:cliffwalk}) the objective is to reach the goal 
state while avoiding the row of terminal ``cliff'' states along the bottom edge 
by controlling four discrete actions up, down, left, right. The state is 
encoded as a binary vector. The environment provides the agent a reward of -1 
at each step and a reward of -100 for entering the cliff. The goal provides no 
reward and terminates the episode. In our experimentation the environment was 
considered solved when the agent achieved a 100 episode moving average reward 
of at least -30.

\begin{figure}[h]
	\centering
	\includegraphics[width=40mm]{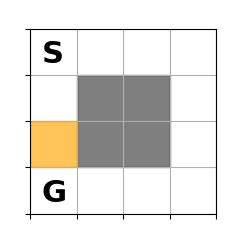}
	\caption{Absent Supervisor Gridworld\label{fig:absent_supervisor}}
\end{figure}

In the ``Absent Supervisor'' gridworld environment (Figure
\ref{fig:absent_supervisor}) the objective is to reach the 
goal state by controlling four discrete actions up, down, left, right. The four 
center squares are impassable. For each episode a supervisor is absent or 
present with uniform probability. The state is encoded as a binary vector. The 
environment provides the agent a reward of -1 at each time step and a reward of 
+50 for entering the goal. When the supervisor is present the orange state, 
located immediately above the goal state, highlighted in Figure 
\ref{fig:absent_supervisor} provides a large negative reward (-30) but no such 
reward when the supervisor is absent. We would like the agent to never pass 
through the orange punishment state. The intent of the environment is to 
demonstrate that when provided the opportunity to cheat by passing through the 
orange state when the supervisor is absent traditional deep reinforcement 
learning algorithms will do so.

\begin{figure}[h]
	\includegraphics{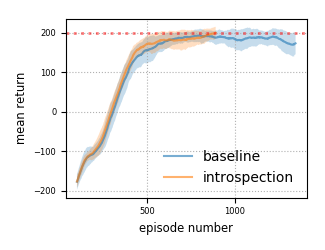}
	\caption{Episodes until ``Lunar Lander'' solved for DDQN (baseline) with 
	and without 
		Introspection Learning\label{fig:results}}
\end{figure}

In each case, the policy $\pi$ considered was a neural network with two hidden layers each
having 32 nodes and hyperbolic tangent activations.  The output
activation was linear with one node for each action. DDQN with soft
target network updates \cite{Lillicrap2015}, the proportional variant of 
Prioritized Experience Replay\cite{Schaul2015}, and an $\epsilon$ greedy 
exploration policy were employed to train the agent 
with the hyperparameters summarized in Table 
\ref{tab:hypers}.

\begin{table}[h]
	\begin{center}
		\begin{tabular}{l|l}
			\textbf{Hyperparameter} & \textbf{Value}\\
			\hline
			experience replay every $n$ timesteps & 2\\
			replay buffer size & 1e5\\
			batch size & 64\\
			$\gamma$ (Discount factor) & 0.99\\
			$\alpha$ (Learning rate) & 1e-3\\
			$\tau$ (Soft target network update rate) & 1e-2\\
			PER $\alpha$ (TD error prioritization) & 0.6\\
			PER $\beta$ (Bias correction) & 0.6\\
		\end{tabular}
		\caption{DDQN hyperparameters used during training\label{tab:hypers}
}
	\end{center}
\end{table}

In the ``Lunar Lander'' environment, the Introspection Learning parameters were 
set as follows.  For the query schedule,
we determine at what interval batches will be searched for and when
searching for batches will cease and training will proceed as normal.
We experimented with solving for state batches at a predetermined interval 
(every 100 episodes) and ceasing when the 100 episode moving average reward 
crossed a predetermined threshold. 
For training on state batches,  states found were treated as terminal
states with high negative reward (-100) 
as determined by the rules of the environment for terminal states.
We have generally found that
incorporating the state batches into the replay buffer 
is beneficial early in the learning process when the policy is poor, as it 
introduces bias (cf. \cite{Schaul2015}).The query 
constraints in both cases were to look for states whose $x$-coordinates were 
outside of the landing zone ($x<-0.25$ or $x> 0.25$), such that the
agent favors selecting an action that would result in it moving
further away from from the landing zone.\footnote{Note
	that alternative choices of query constraints are also possible
	including, e.g., querying for those states that move the agent in the
	correct direction, which could be given extra reward.  Our approach
	here is based on trying to minimize the number of obviously risky
	actions the agent is likely to carry out during training, while
	allowing the agent freedom to explore reasonable actions.}
      
This region of the state-space was divided into boxes using a
simple quantization scheme that
ignored regions of state space where examples satisfying the query
constraints would be impossible to find.  In general, such
quantization schemes  should be
sufficiently fine-grained to allow generation of many and diverse
examples.  Twenty training runs 
with a set of twenty random seeds were run with and without our approach for a 
maximum of 500,000 timesteps. Results averaged over the training runs are 
summarized in Figure \ref{fig:results}.  DDQN with Introspection Learning 
solved the environment in a mean of 893 episodes while DDQN without 
Introspection Learning (baseline) failed to successfully solve the environment 
on average within 500,000 timesteps. 

\begin{figure}[h]
	\includegraphics[scale=0.5]{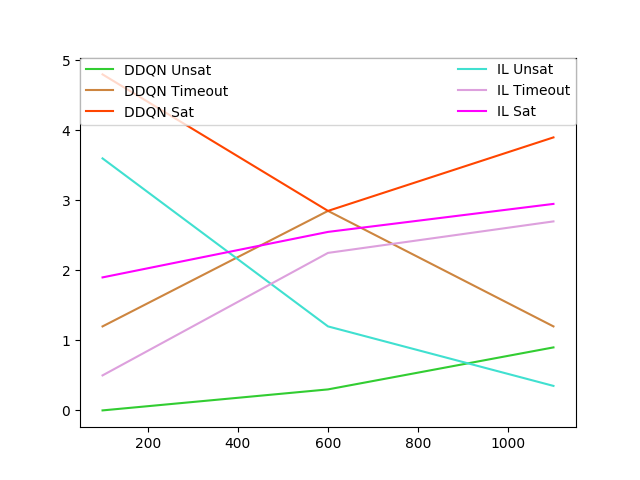}
	\caption{Total number of unsat instances as a function of
		time for baseline (DDQN) and IL.\label{fig:unsats}
}
\end{figure}

In addition to observing performance benefits, we also evaluated the
agents trained with Introspection Learning for robustness benefits.
In particular, we periodically stored the weights of both the
Introspection Learning agent and the baseline agent during training
for each of the twenty runs.  We then recorded, for different regions
of state space, statistics regarding the Sat, Unsat and Timeout results obtained
when querying the SMT-solver on these agents across training.  To
recall, in this case, a Sat result indicates that there exists a state
$s$ in the specified region $U$ of state space such that an undesirable action
$\pi(s)$ (in this case, moving away from the landing zone) is selected by the
agent.  Likewise, an Unsat result indicates that there
is a mathematical proof that there exists no state $s$ in $U$ such
that $\pi(s)$ is undesirable.  We gathered Sat, Unsat and Timeout data
across a number of different selections of $U$.  Tables
\ref{tab:small_baseline} and \ref{tab:small_il} record the percentages
of each kind of result across all twenty test runs that were captured
at four points during training. The selection of $(U_{i})_{i}$
queried here were a subset of the subsets of (state,action)-space
queried during the actual Introspection Learning training and the
results show a clear improvement of robustness over the baseline.
Timeouts during training were set to five seconds and to ten seconds
during evaluation.
One interesting point that we noticed in analyzing the robustness
evaluation data is that larger numbers of Unsat results for the
Introspection Learning agents were obtained at the beginning of
training than the end.  This is illustrated, for a typical example
(the run with ID number 480951) in Figure \ref{fig:unsats}.  This
is likely due to the schedule employed as part of the introspection
learning algorithm and highlights the more general fact that
reinforcement learning agents are sometimes subject to ``forgetting''
important learned behavior at later stages of training. Since the
agents at the end of training were typically very good at solving the
task, the regions of state space in which this forgetfulness would
manifest themselves were likely off-trajectory (i.e., unreachable by
the current policy).

In order to emphasize that this improvement is very much a function of
the specific $(U_{i})_{i}$ used during training, and tested at
evaluation time, we include for comparison in Table \ref{tab:large_percentages} the average
percentages for an alternative selection of $(U_{i})_{i}$ used at
evaluation time.  Here the improvements are more modest.

\begin{table}[h]
	\centering
	\begin{tabular}[h]{l|ccc}
          \textbf{Run ID} & \textbf{Unsat} & \textbf{Sat} & \textbf{Timeout}\\
          \hline
          34001    & 0\% &  62.5\%  &  37.5\%\\
          390797 &  0\% &  100\% &  0\%\\
          747524 & 0\% & 75\% & 25\%\\
          480621 & 25\% & 50\% & 25\%\\
          475982 & 50\% & 25\% & 25\%\\
          319324 & 25\% & 62.5\% & 12.5\%\\
          449374 & 0\% & 50\% & 50\%\\
          491386 & 0\% & 50\% & 50\%\\
          532333 & 0\% & 50\% & 50\%\\
          55487 & 0\% & 75\% & 25\%\\
          4211 & 0\% & 50\% & 50\%\\
          480951 & 0\% & 100\%  & 0\%\\
          219015 & 0\% & 87.5\% & 12.5\%\\
          481614 & 0\% & 75\% & 25\%\\
          367249 & 25\% & 50\% & 25\%\\
          508732 & 0\% & 100\% & 0\%\\
          521233 & 0\% & 50\% & 50\%\\
          543696 & 0\% & 75\% & 25\%\\
          998982 & 0\% & 100\% &  0\%\\
          36067 & 0\% & 75\% & 25\%\\
          \hline
          Average & 6.250\% & 68.125\% & 25.625\%
	\end{tabular}
	\caption{Percentages of Sat, Unsat and Timeout instances
          for Baseline DDQN at four points during training.\label{tab:small_baseline}}
\end{table}
\begin{table}[h]
	\centering
	\begin{tabular}[h]{l|ccc}
          \textbf{Run ID} & \textbf{Unsat} & \textbf{Sat} & \textbf{Timeout} \\
          \hline
          34001 & 50\% & 25\% & 25\%\\
          390797 & 25\% & 25\% & 50\%\\
          747524 & 0\% & 50\% & 50\%\\
          480621 & 25\% & 0\% & 75\%\\
          475982 & 25\% & 50\% & 25\%\\
          319324 & 0\% & 75\% & 25\%\\
          449374 & 0\% & 50\% & 50\%\\
          491386 & 0\% & 45.8333\% & 54.1667\%\\
          532333 & 25\% & 25\% & 50\%\\
          55487 & 25\% & 50\% & 25\%\\
          4211 & 50\% & 25\% & 25\%\\
          480951 & 0\% & 75\% & 25\%\\
          219015 & 25\% & 37.50\% & 37.50\%\\
          481614 & 25\% & 75\% & 0\%\\
          367249 & 25\% & 75\% & 0\%\\
          508732 & 0\% & 75\% & 25\%\\
          521233 & 50\% & 50\% & 0\%\\
          543696 & 79.1667\% & 0\% & 20.8333\%\\
          998982 & 25\% & 33.3333\% & 41.6667\%\\
          36067 & 25\% & 0\% & 75\%\\
          \hline
          Average & 23.958\% & 42.083\% & 33.958\%
	\end{tabular}
	\caption{Percentages of Sat, Unsat and Timeout instances
          for Introspection Learning at four points during training.\label{tab:small_il}}
\end{table}
\begin{table}[h]
	\centering
	\begin{tabular}[h]{l|ccc}
          \textbf{Run ID} & \textbf{Unsat} & \textbf{Sat} & \textbf{Timeout}\\
          \hline
          \textbf{Baseline} & 83.3\% & 1.4\% & 15.3\%\\
          \textbf{Introspection} & 85.3\% & 0.6\% & 14.2\%
	\end{tabular}
	\caption{Average percentages of Sat, Unsat and Timeout instances
          for Baseline DDQN versus Introspection Learning for the
          full batch of all twenty runs on another selection of query
          subsets $(U_{i})_{i}$.  For this choice of subsets, the
          gains in robustness are more modest.\label{tab:large_percentages}}
\end{table}

\begin{figure}[h]
	\centering
	\includegraphics[width=40mm]{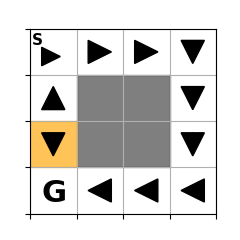}
	\caption{Final DDQN policy with Introspection Learning does not select to 
	enter the orange punishment state when the supervisor is absent.\label{fig:absent_supervisor_introspection}}
\end{figure}

\begin{figure}[h]
	\centering
	\includegraphics[width=40mm]{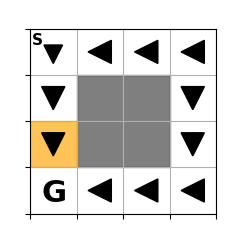}
	\caption{Final DDQN policy without Introspection Learning (baseline) 
		selects to cheat and enter the orange punishment state when the 
		supervisor is absent.\label{fig:absent_supervisor_baseline}}
\end{figure}

In the ``Absent Supervisor'' environment the Introspection Learning parameters 
were set as follows. Solving for state batches is unnecessary as in this 
discrete state environment we are only concerned with the agent choosing to 
enter the orange punishment state from the state directly above it. For the 
query schedule solving for this specific behavior is performed at every 
timestep and during training this transition is treated as terminal with high 
negative reward (-100). Results for DDQN with and without Introspection 
Learning are provided in Figures \ref{fig:absent_supervisor_introspection} and 
\ref{fig:absent_supervisor_baseline} respectively.  One interesting
point about the ``Absent Supervisor'' environment is that, for the
evident notion of good policy, one of the hypotheses (the ``Strong
Compatiblity'' assumption) of our Theorem \ref{theorem:equiv} is violated.

\begin{figure}[h]
	\centering
	\includegraphics[width=85mm]{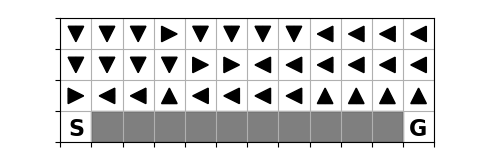}
	\caption{Representative DDQN policy with Introspection Learning after 30 
	episodes of training has learned a safer policy of avoiding the cliff.\label{fig:cliffwalk_introspection_30}}
\end{figure}

\begin{figure}[h]
	\centering
	\includegraphics[width=85mm]{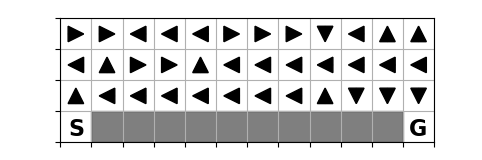}
	\caption{Representative DDQN policy without Introspection Learning 
	(baseline) after 30 training episodes still selects to enter the cliff from 
	some states.\label{fig:cliffwalk_baseline_30}}
\end{figure}

In the ``Cliff Walk'' environment the Introspection Learning parameters 
were set as follows. Solving for state batches is unnecessary as in this 
discrete state environment we are only concerned with the agent choosing to 
enter the cliff states which can only be done from the state directly above 
each cliff state respectively. For the 
query schedule solving for these specific behaviors is performed at every 
timestep and during training this transition is treated as terminal with high 
negative reward (-100). It should be noted that in this particular case the 
environment already treats these transitions as terminal with high negative 
reward (-100) and thus Introspection Learning will not alter the optimal 
policy (in particular, the hypotheses of Theorem \ref{theorem:equiv}
are satisfied). 
In this experiment, five training runs with a set of five random seeds were run with and without 
our approach until the environment was solved. During training, at each 
timestep, a running count was kept of the number of states from which the agent 
would 
select to enter the cliff states ``lemming''. During training the policies were 
found to lemming on average 112 times with Introspection Learning and 29,501 
times without. It was experimentally found that an agent with Introspection 
Learning would rarely learn a policy during training that would enter the cliff 
after the first training episode while it was routine for an agent without 
Introspection Learning. Representative policies learned by DDQN with and 
without Introspection Learning after 30 training episodes are provided in 
Figures \ref{fig:cliffwalk_introspection_30} and 
\ref{fig:cliffwalk_baseline_30} respectively. Additionally, agents with 
Introspection Learning enjoyed a small performance benefit solving the 
environment in 208 episodes on average over the five training runs 
while agents without Introspection Learning averaged 229 episodes to solve the 
environment.

\section{Conclusions}

In this paper we have introduced a novel reinforcement learning
algorithm based on ideas coming from formal methods and SMT-solving.
We have shown that, on suitable problems, these techniques can be
employed in order to improve robustness of RL agents and to speed up
their training.  We have also given examples of how SMT-solving can be
used to analyze reinforcement learning agent robustness.  There are a
number of extensions of this preliminary work possible.  We mention
several prominent directions here.

First, the focus here has been on single-step analysis of agent
behavior, but a reachability analysis approach focused on
trajectories leading to target states would likely generate more
relevant data for learning.  E.g., consider a geo-fenced space that we
do not want the agent to enter and that is reachable through many different
(state, action) combinations.  Once a violation occurs, we would like
to examine the trajectory in order to learn what earlier choices led
the agent there.

Second, whereas in our ``lunar lander'' experiments we utilized an
\emph{ad hoc} quantization of the state space, it should be in many
cases possible to learn such regions as part of the algorithm.  This
is a hard search problem so relying on these parameterizations is
necessary and should therefore be automated.
In conjunction with the reachability analysis mentioned above, this
approach is likely to give more targeted and therefore useful data to
include in the replay buffer.

Finally, while the SMT-solving technology being used is sufficient for
low-dimensional state-spaces, these techniques face
scalability issues on large state-spaces such as those coming from
video data.  How to handle these higher-dimensional state-spaces in a
similar way is one of the exciting challenges in this area.

\subsection{Acknowledgments}

We would like to thank Ramesh S, Doug Stuart, Huafeng Yu, Sicun Gao,
Aleksey Nogin, and Pape Sylla for useful conversations on topics
related to this paper.  We are also grateful to Tom Bui, Bala Chidambaram, Cem
Saraydar, Roy Matic, Mike Daily and Son Dao for their support of and
guidance regarding this research.  Finally, we would like to thank
Alessio Lomuscio and Clark Barrett for their interest in this work and
for encouraging us to capture these results in a paper.

\section{Appendix: Theoretical Results}\label{appendix}

Fix a pre-MDP $D$ and assume given a (non-empty) subset $\goods$ of
$\policies(D)$ which we regard as the \emph{good} policies: those
policies $\pi$ whose $(s,\pi(s))$ have the properties of interest.
\begin{definition}\label{def:mdp_equiv}
  MDPs $M$ and $M'$ are \emph{equivalent} whenever
  $\optimals(M)\subseteq\optimals(M')$ and $\optimals(M')\subseteq\optimals(M)$.
\end{definition}
Furthermore, throughout this section we assume given a \emph{fixed} MDP $M=(r,T)$ in
$\mdps(D)$.  Additionally, assume given a fixed discount factor
$0\leq\gamma < 1$.  We also adopt throughout this section two further
hypotheses, which we now describe.
\begin{assumption}[Bad Set]\label{assumption:bad_acts}
  There exists a subset $B\subseteq S\times A$ such that $\pi$ is in
  $\goods$ if and only if, for all $(s,a)\in B$, $\pi(s)\neq a$.
\end{assumption}
Our next hypothesis guarantees that the reward structure is already
sufficiently compatible with $\goods$.
\begin{assumption}[Strong Compatibility]\label{assumption:sc}
  All optimal policies for $M$ are in $\goods$.  I.e., $\optimals(M)\subseteq\goods$.
\end{assumption}

We define a new MDP structure
$M^{\dagger}=(r^{\dagger},T^{\dagger})$ in $\mdps(D)$ by
\begin{align*}
  T^{\dagger} & := T\cup B,\text{ and}\\
  r^{\dagger}(s,a) & :=
                     \begin{cases}
                       -1 + \min_{\pi}Q^{\pi}_{M}(s,a) & \text{ if
                       }(s,a)\in B\text{, and}\\
                       r(s,a) & \text{ otherwise.}
                     \end{cases}
\end{align*}
It is straightforward to prove that $r^{\dagger}$ is bounded since $r$
is.  Note that we are also modifying the underlying pre-MDP here by now
imposing the condition that $a_{b}\colon s_{b}\to s_{t}$.

An immediate proof of the following proposition can be obtained using
the
notion of \emph{bounded corecursive algebra} from \cite{Moss}, where it is shown that the state-value functions
$V^{\pi}_{M}\colon S\to\mathbb{R}$ are canonically determined by the
generating maps $v^{\pi}_{M}\colon S\to \mathbb{R}\times D(S)$ given
by
\begin{align*}
  v^{\pi}_{M}(s) & := (r(s,\pi(s)),p(s,\pi(s))),
\end{align*}
where $D(-)$ is the probability distribution monad.
\begin{proposition}\label{proposition:sc_V}
  If $\pi$ is in $\goods$, then $V^{\pi}_{M^{\dagger}}=V^{\pi}_{M}$.
  \begin{proof}
    It suffices to show that $v^{\pi}_{M^{\dagger}}=v^{\pi}_{M}$, which is trivial for $\pi$ in $\goods$.
  \end{proof}
\end{proposition}
\begin{corollary}\label{cor:sc_Q}
  If $\pi$ is in $\goods$, then
  $Q^{\pi}_{M^{\dagger}}(s,a)=Q^{\pi}_{M}(s,a)$ if and only if
  $(s,a)\notin B$.
\end{corollary}
\begin{lemma}\label{lemma:sc_ineq}
  $\optimals(M)\subseteq\optimals(M^{\dagger})$.
  \begin{proof}
    Suppose given an optimal policy $\pi$ for $M$.  By Bellman
    optimality, $\pi$ is optimal for $M^{\dagger}$ if and
    only if, for all $s$,
    \begin{align*}
      \pi(s)\in\argmax_{a}Q^{\pi}_{M^{\dagger}}(s,a).
    \end{align*}
    Let $s$ and $a$ be given.  There are two cases
    depending on whether or not $(s,a)\in B$.

    When $(s,a)\notin B$,
    \begin{align*}
      Q^{\pi}_{M^{\dagger}}(s,a) & = Q^{\pi}_{M}(s,a) \\
       & \leq  Q^{\pi}_{M}(s,\pi(s))\\
      & =  Q^{\pi}_{M^{\dagger}}(s,\pi(s)),
    \end{align*}
    where the equations are by Corollary \ref{cor:sc_Q} and the
    inequality is by optimality of $\pi$.

    When $(s,a)\in B$,
    \begin{align*}
      Q^{\pi}_{M^{\dagger}}(s,a)
      & = -1 + \min_{\pi'}Q^{\pi'}_{M}(s,a) + 0\\
      & \leq -1 + Q^{\pi}_{M}(s, a)\\
      & < Q^{\pi}_{M}(s, a)\\
      & \leq Q^{\pi}_{M}(s,\pi(s))\\
      & = Q^{\pi}_{M^{\dagger}}(s,\pi(s)),
    \end{align*}
    where the final inequality is by optimality of $\pi$ and the final
    equality is by Corollary \ref{cor:sc_Q}.
  \end{proof}
\end{lemma}
\begin{lemma}\label{lemma:dagger_good}
  $\optimals(M^{\dagger})\subseteq\goods$.
  \begin{proof}
    Let a policy $\pi$ for $M^{\dagger}$ be given such that, for some
    $s$, $(s,\pi(s))\in B$ and let $\pi'$ be an optimal policy for $M$.
    Then
    \begin{align*}
      Q^{\pi}_{M^{\dagger}}(s,\pi(s))  & < Q^{\pi}_{M}(s,\pi(s))\\
                                       &\leq Q^{\pi'}_{M}(s,\pi'(s)) \\
                                       & = Q^{\pi'}_{M^{\dagger}}(s,\pi'(s)),
    \end{align*}
    so that such a $\pi$ cannot be optimal.
  \end{proof}
\end{lemma}
\begin{theorem}\label{theorem:equiv}
  $M$ and $M^{\dagger}$ are equivalent.
  \begin{proof}
    By Lemma \ref{lemma:sc_ineq} it suffices to show that
    $\optimals(M^{\dagger})\subseteq\optimals(M)$, which is immediate since
    \begin{align*}
      V^{\pi}_{M} = V^{\pi}_{M^{\dagger}} = V^{\pi'}_{M^{\dagger}} = V^{\pi'}_{M},
    \end{align*}
    for any optimal policy $\pi$ for $M^{\dagger}$ and any optimal policy
    $\pi'$ for $M$.  Here the first equation is by Proposition
    \ref{proposition:sc_V} and Lemma \ref{lemma:dagger_good}, the second
    equation is by optimality of $\pi'$ for $M^{\dagger}$ by Lemma
    \ref{lemma:sc_ineq}, and the final equation is by Proposition
    \ref{proposition:sc_V} and the Strong Compatibility hypothesis.
  \end{proof}
\end{theorem}

\printbibliography

\end{document}